\documentclass[11pt]{article}
\usepackage{eacl2017}
\usepackage{times}
\usepackage{url}
\usepackage{latexsym}
\usepackage{graphicx}
\usepackage{multirow}
\usepackage{amsmath}

\eaclfinalcopy % Uncomment this line for the final submission
\def\eaclpaperid{11} %  Enter the acl Paper ID here

\title{Exploring Different Dimensions of Attention for Uncertainty Detection}

\author{Heike Adel \and  Hinrich Sch\"{u}tze \\
Center for Information and Language Processing (CIS) \\ 
       LMU Munich, Germany\\
  {\tt heike@cis.lmu.de} \\}

\date{}

\def\figref#1{Figure~\ref{fig:#1}}
\def\figlabel#1{\label{fig:#1}\label{p:#1}}

\def\tabref#1{Table~\ref{tab:#1}}
\def\tablabel#1{\label{tab:#1}\label{p:#1}}

\def\secref#1{Section~\ref{sec:#1}}
\def\seclabel#1{\label{sec:#1}\label{p:#1}}
\def\eqref#1{Eq.~\ref{eqn:#1}}

\def\eqlabel#1{\label{eqn:#1}}

\begin{document}

\maketitle

\begin{abstract} 
Neural networks with attention have proven effective
for many natural language processing tasks. In this paper, 
we develop attention mechanisms for uncertainty detection.
In particular, we
generalize standardly used attention mechanisms
by introducing \emph{external attention}
and \emph{sequence-preserving attention}. These novel architectures 
differ from standard approaches in that they use external
resources to compute attention weights and preserve sequence
information. We compare them 
to other configurations along different dimensions of attention. 
Our novel architectures set the new state 
of the art on a Wikipedia benchmark dataset and perform
similar to the state-of-the-art model
on a biomedical benchmark which uses
a large set of linguistic
features.
\end{abstract}

\section{Introduction}
\seclabel{intro}
For many natural language processing (NLP) tasks,
it is essential to
distinguish uncertain (non-factual) from certain (factual)
information.  Such tasks include information extraction,
question answering, medical information
retrieval, opinion detection, sentiment analysis
\cite{karttunen,vincze,cruz2015} and 
knowledge base
population (KBP).
In KBP,  we need to distinguish, e.g.,
``X may be Basque'' and ``X was rumored to be Basque''
(uncertain) from ``X is Basque'' (certain)
to decide whether to add the fact ``Basque(X)'' to a
knowledge base.
In this paper, we use the term \emph{uncertain
 information} to refer to speculation, opinion, vagueness
and ambiguity.
We focus our experiments on the 
uncertainty detection (UD) dataset from the
CoNLL2010 hedge cue detection
task \cite{CoNLLsharedTask}. It consists of two medium-sized 
corpora from different domains (Wikipedia and biomedical)
that allow us to run a large number of
comparative experiments with different neural networks 
and exhaustively investigate
different dimensions of attention.

Convolutional  and
recurrent neural networks (CNNs and RNNs)
perform well on many
NLP tasks
\cite{cw,kalchbr,zeng2014,rnnRelClass}.
CNNs are most often used with pooling.
More recently, attention mechanisms have been 
successfully integrated into CNNs and RNNs
\cite{attention0,attention1,attention2,attention3,attention5,attention6,abcnn}. Both
pooling and attention can be thought of as \emph{selection
mechanisms} that help the network focus on the most relevant
parts of a layer, either an input or a hidden layer.
This is especially beneficial for long input sequences, e.g.,
long sentences or entire documents.
We apply CNNs and RNNs to uncertainty detection and compare them to a number of baselines.
We show that attention-based CNNs and RNNs are
effective for uncertainty detection.
On a Wikipedia benchmark, we improve the state of the art by
more than 3.5 $F_1$ points. 

Despite the success of attention in prior work, the design
space of related network architectures has not been fully
explored. In this paper, we develop novel ways to 
calculate attention weights and integrate them
into neural networks. Our models are motivated
by the characteristics of the uncertainty task,
yet they are also a first attempt to systematize
the design space of attention.
In this paper, we begin with investigating three dimensions
of this space: weighted vs. unweighted selection, 
sequence-agnostic vs. sequence-preserving selection,
and internal vs. external attention.

\def\myfigheight{.18}

\begin{figure*}
\begin{tabular}{c@{\hspace{0cm}}c@{\hspace{0cm}}c@{\hspace{0cm}}c}
(1) & (2) & (3) & (4)\\
  \includegraphics[height=\myfigheight\textheight]{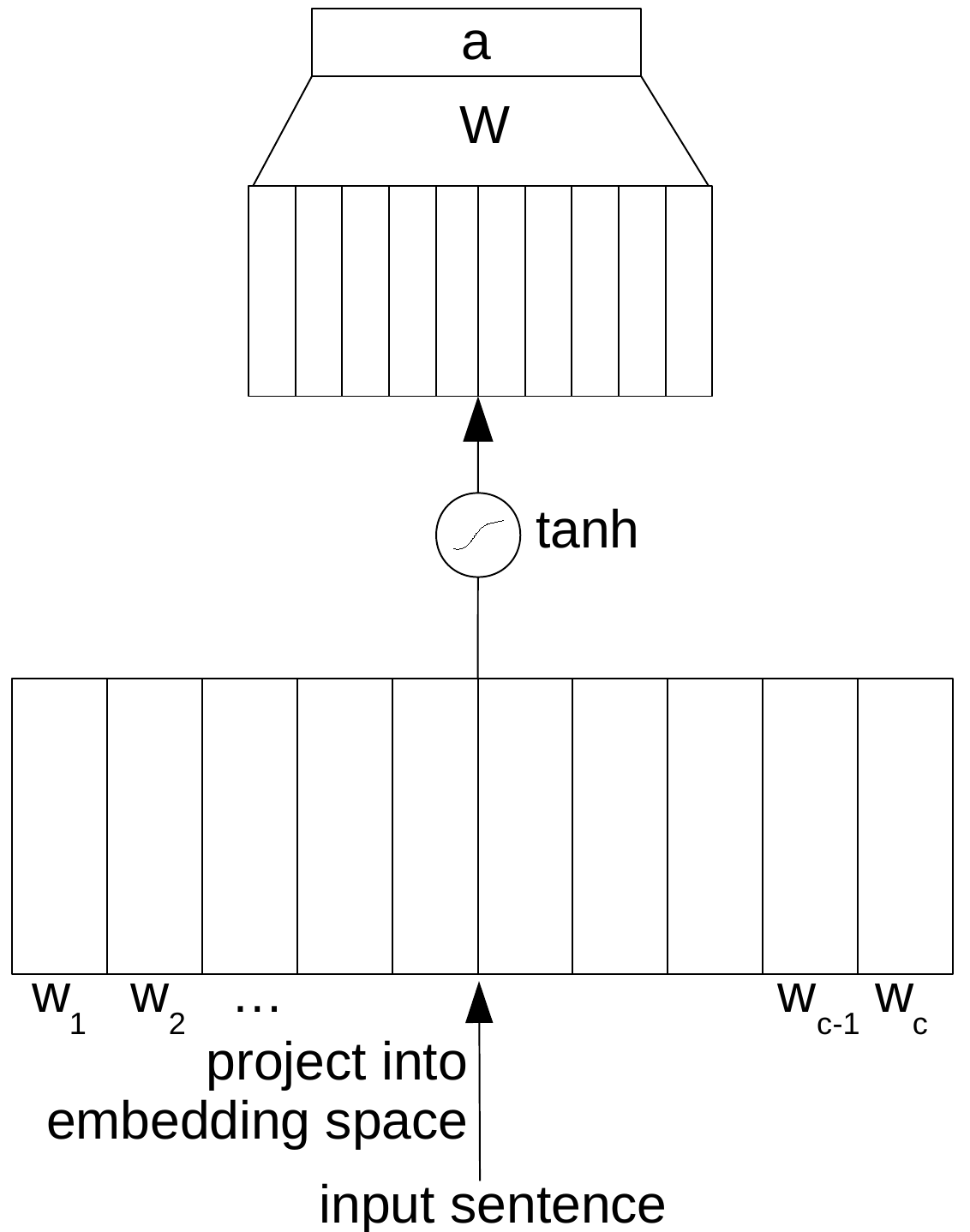}&
 \includegraphics[height=\myfigheight\textheight]{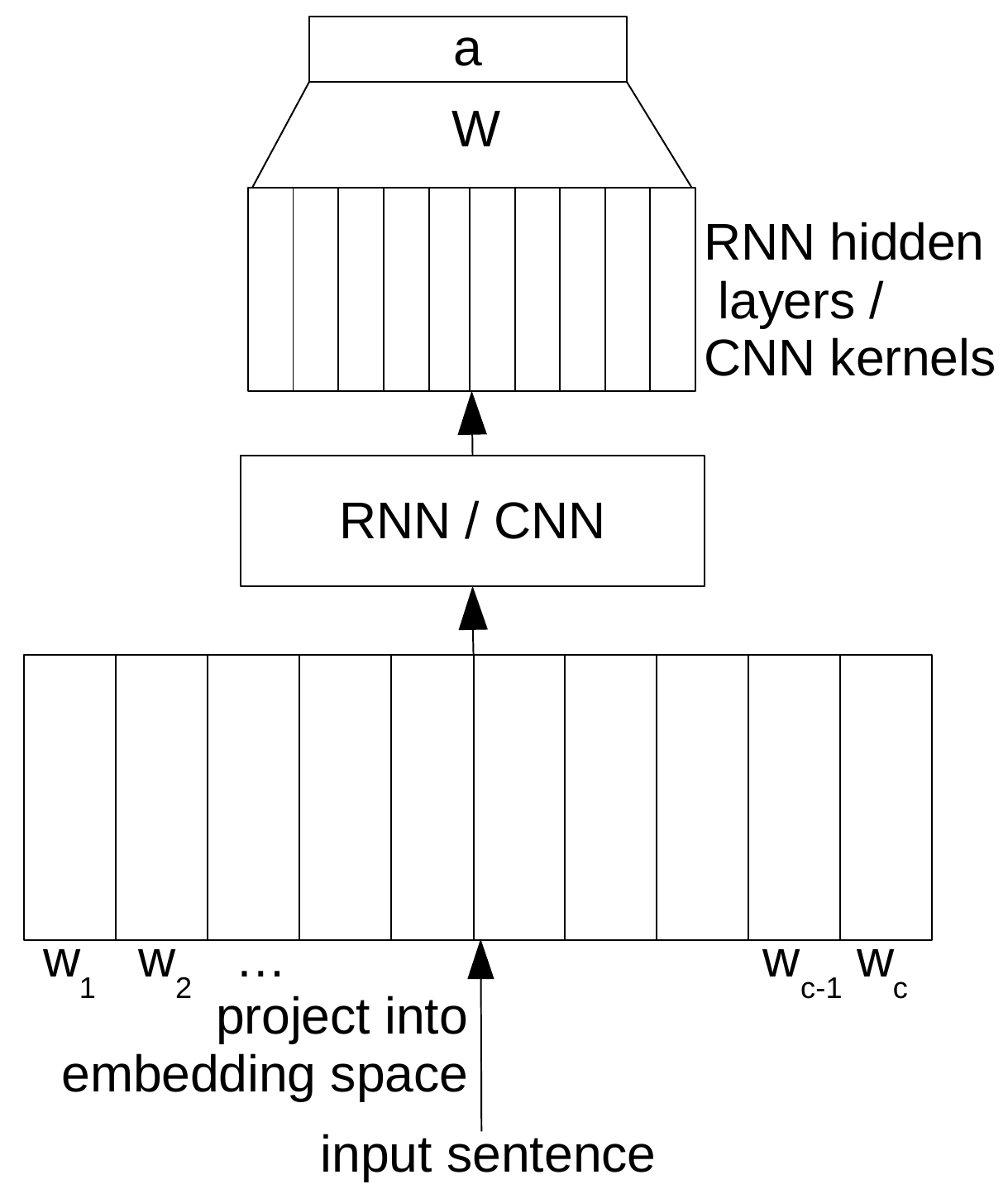}&
 \includegraphics[height=\myfigheight\textheight]{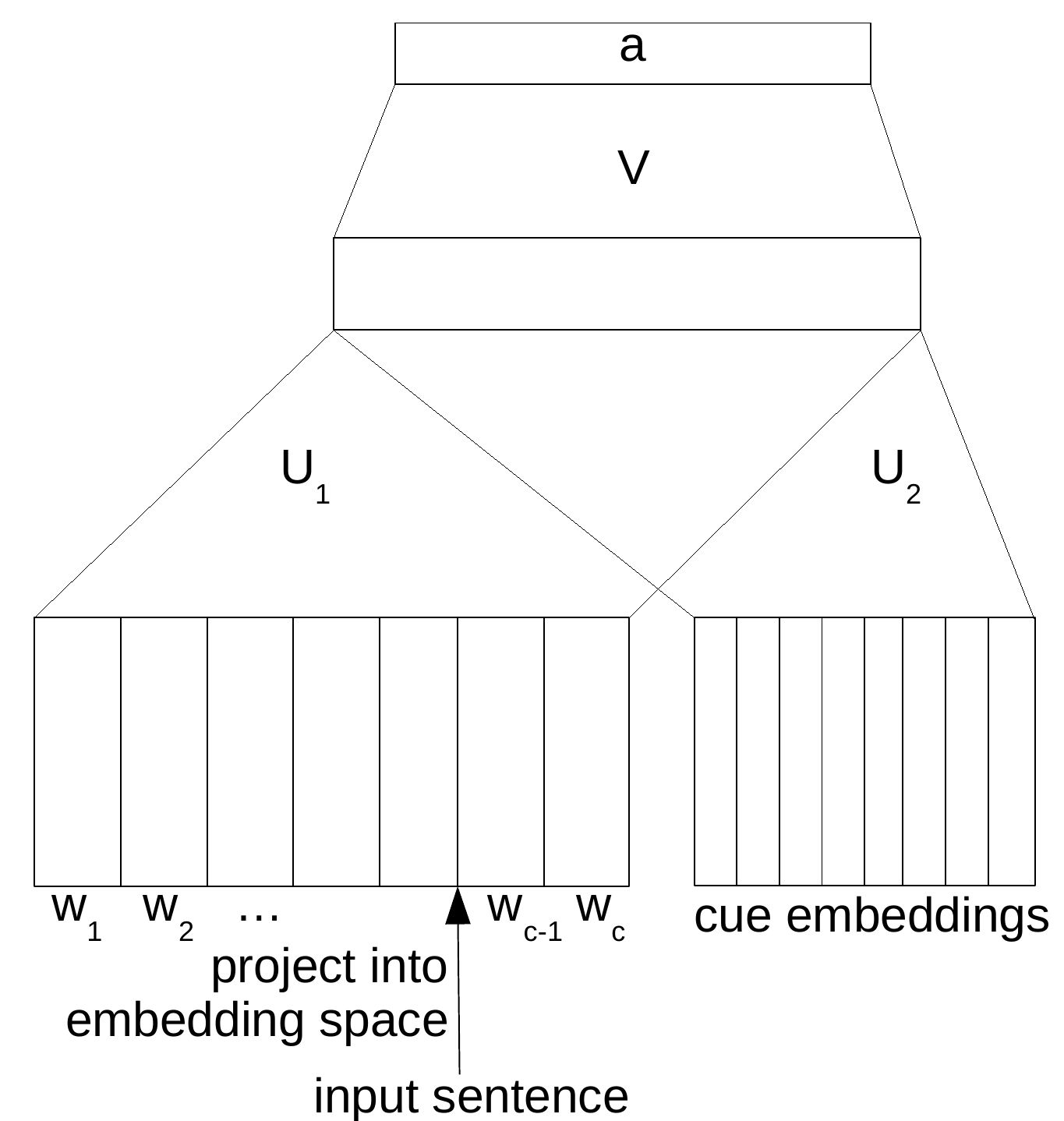}&
 \includegraphics[height=\myfigheight\textheight]{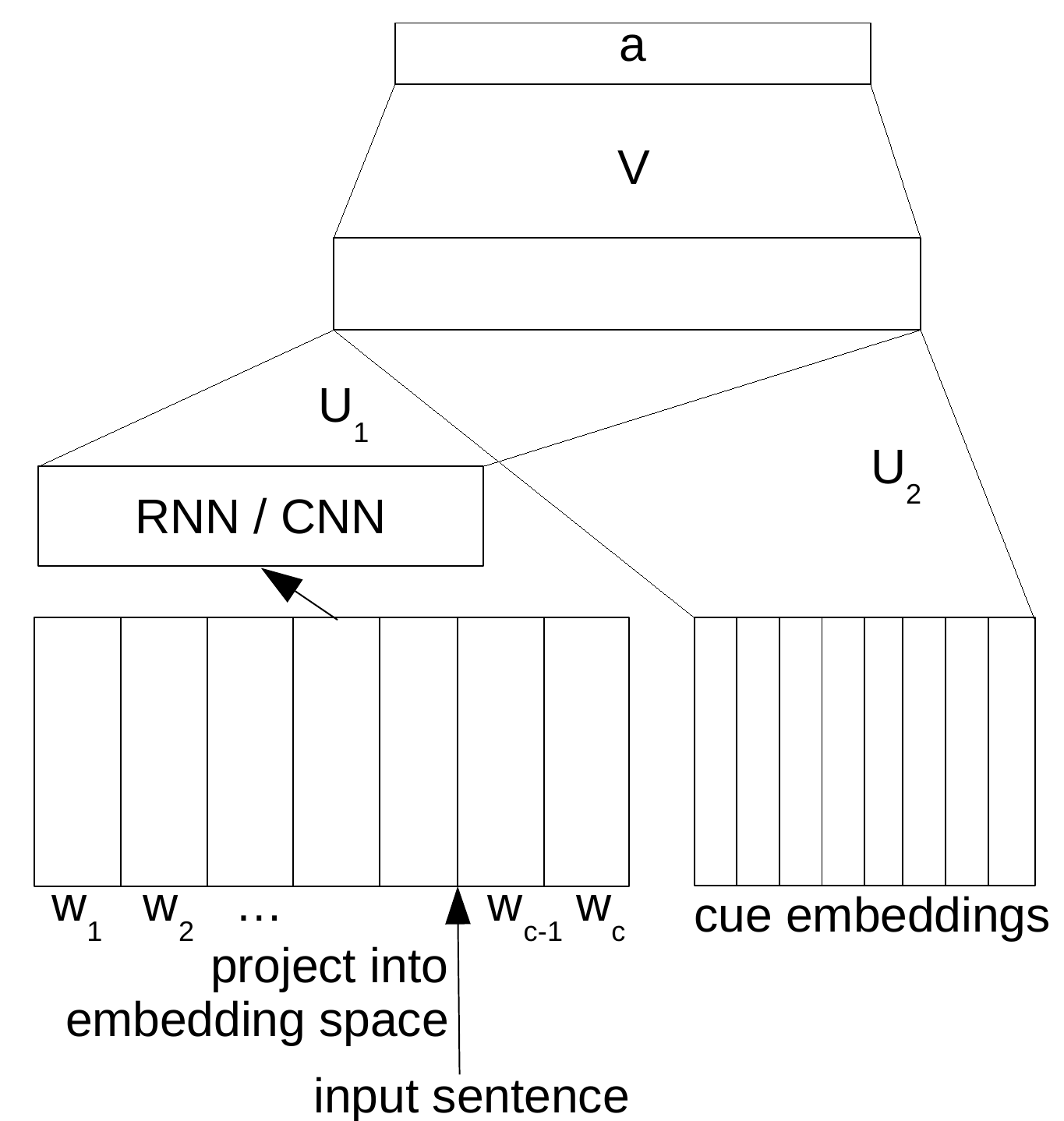}
\end{tabular}
 \caption{Internal attention on (1)  input and (2) hidden
   representation. External attention on 
   (3)  input and (4)  hidden representation. For the whole network structure, see \figref{wholeNet}.}
 \figlabel{fig-attention}
\end{figure*}

\textbf{Weighted vs.\ Unweighted Selection.} Pooling is
unweighted selection: it outputs the
selected values as is. In contrast, attention can be thought
of as weighted selection: some input elements are highly
weighted, others receive weights close to zero and are
thereby effectively not selected. The advantage of weighted
selection is that the model learns to decide based on the
input how many values it should select. Pooling either selects
all values (average pooling) or k values (k-max
pooling). If there are more than k uncertainty cues in
a sentence, pooling is not able to focus on all of them.

\textbf{Sequence-agnostic vs.\ Sequence-preserving
Selection.} K-max pooling \cite{kalchbr} is sequence-preserving: it
takes a long sequence as input and
outputs a subsequence whose members are in the same order as
in the original sequence. In contrast, attention is
generally implemented as a weighted average of the input
vectors. That means that all ordering information is lost
and cannot be recovered by the next layer.
As an alternative, we present and evaluate new 
sequence-preserving ways of attention. For uncertainty detection,
this might help distinguishing phrases like
``it is not uncertain that X is Basque'' and 
``it is uncertain that X is not Basque''.

\textbf{Internal vs.\ External Attention.} 
Prior work calculates attention weights based on the input
or hidden layers of the neural network. We call this internal
attention. For uncertainty detection, it can be
beneficial to give the model a lexicon of seed
cue words or phrases. 
Thus, we provide the network with additional
information to bear on identifying and 
summarizing features.
This can simplify the training process by
guiding the model to recognizing
uncertainty cues. 
We call this external attention and show 
that it improves performance for uncertainty detection.

Previous work on attention and pooling has only
considered a small number of the possible configurations 
along those dimensions of attention. 
However, the internal/external  and
un/weighted distinctions can potentially
impact performance because external resources add information
that can be critical for good performance and because weighting
increases the flexibility and expressivity of neural network models.
Also, word order is often critical for meaning
and is therefore an important feature in NLP. 
Although our models are motivated by the characteristics
of uncertainty detection, they could be useful for other NLP tasks as well.

Our main contributions are as follows.  
(i) We extend the design space of selection mechanisms for
neural networks and conduct an extensive set of experiments testing
various configurations along several
dimensions of that space, including novel sequence-preserving
and external
attention mechanisms.
(ii) To our knowledge, we are the first to apply 
convolutional and recurrent
neural networks to uncertainty detection.
We demonstrate the effectiveness of the proposed 
attention architectures for this task and set 
the new state of the art
on a Wikipedia benchmark dataset.
(iii) We publicly release our code for future research.\footnote{\url{http://cistern.cis.lmu.de}}

\section{Models}
\textbf{Convolutional Neural Networks.}
CNNs have been successful for many NLP tasks since 
convolution and pooling can detect key features independent
of their position in the sentence. Moreover, they
can take advantage of word embeddings and their characteristics. Both
properties are also essential for uncertainty detection
since we need to detect cue phrases that can occur anywhere in
the sentence; and since some notion of similarity improves
performance if a cue phrase in the test data did not occur in the
training data, but is similar to one that did. The CNN we
use in this paper has one convolutional layer, 3-max pooling
(see \newcite{kalchbr}), a fully connected hidden layer and
a logistic output unit.

\textbf{Recurrent Neural Networks.}
Different types of RNNs
have been applied widely to NLP tasks, including language modeling \cite{rnnlm01,rnnlm02},
machine translation \cite{gatedRNN,attention0}, relation classification \cite{rnnRelClass} and entailment
\cite{attention3}.
In this paper, we apply a bi-directional gated RNN (GRU)
with gradient clipping and a logistic output unit.
\newcite{chung} showed that GRUs and LSTMs have similar
performance, but GRUs are more efficient in training.
The hidden layer $h$ of the GRU is 
parameterized by two matrices $W$ and $U$
and four additional matrices
$W_r$, $U_r$ and $W_z$, $U_z$ for the reset gate $r$ and 
the update gate $z$ \cite{gatedRNN}:
\begin{eqnarray}
 r = \sigma(W_rx + U_rh^{t-1})\\
 z = \sigma(W_zx + U_zh^{t-1})\\
 h^t = z \odot h^{t-1} + (1-z)\odot \tilde{h}^{t}\\
 \tilde{h}^t = \sigma(Wx + U(r \odot h^{t-1})) \eqlabel{reset}
\end{eqnarray}
$t$ is the index for the current time step,
$\odot$ is  element-wise multiplication and
$\sigma$ is the sigmoid.

\section{Attention}
\seclabel{gatedCNN}
\subsection{Architecture of the Attention Layer}
We first define
an attention layer $a$ for input $x$:
\begin{eqnarray}
\alpha_i &= &\frac{\exp(f(x_i))}{\sum_j \exp(f(x_j))} \eqlabel{gating2}\eqlabel{attweight}\\
a_i &= & \alpha_i \cdot x_i\eqlabel{gating1}\eqlabel{attreweighted}
\end{eqnarray}
where $f$ is a scoring function,
the $\alpha_i$ are the attention weights and each input $x_i$ is
reweighted by its corresponding attention weight $\alpha_i$.

The most basic definition of $f$ is as a linear  scoring
function on the input $x$:
\begin{equation}
f(x_i) = W^Tx_i \eqlabel{internalatt}
\end{equation}
$W$ are parameters that are learned in training.

\begin{figure}
 \centering
 \includegraphics[height=.08\textheight]{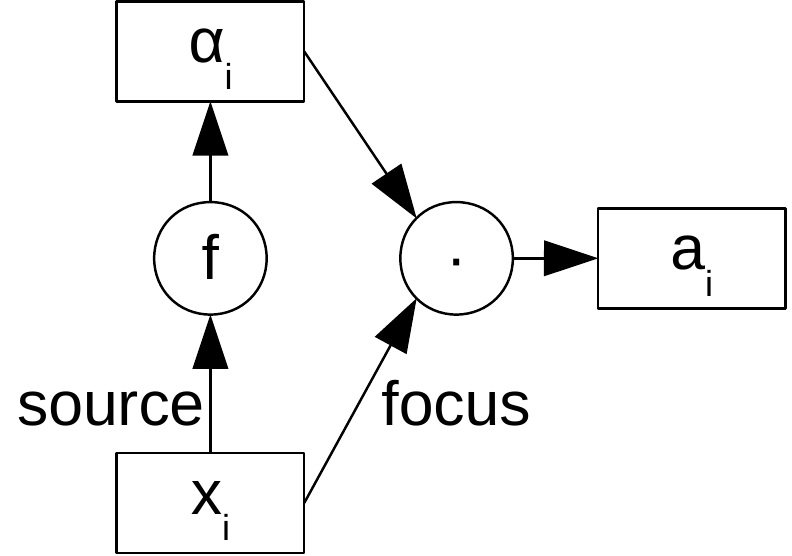}
 \includegraphics[height=.08\textheight]{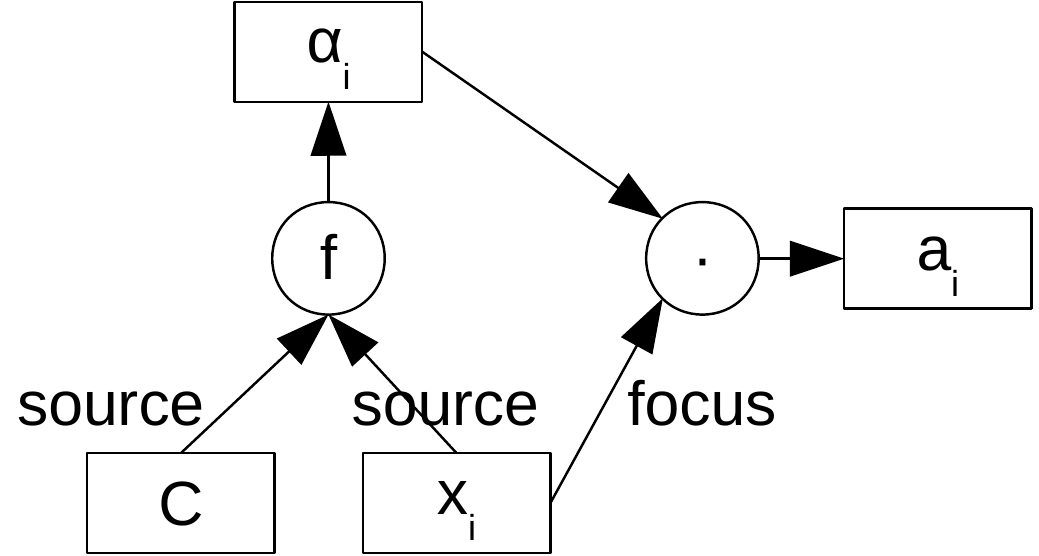}
 \caption{Schemes of focus and source: left: internal attention, right: external attention}
 \figlabel{schemes}
\end{figure}

\subsection{Focus and Source of Attention}
In this paper, we distinguish between focus and source of
attention. 

The \emph{focus} of attention is the
layer of the network that is reweighted by attention
weights, corresponding to $x$ in \eqref{attreweighted}. 
We consider two
options for the application in uncertainty detection as
shown in \figref{fig-attention}: (i) the focus is on
the input, i.e., the matrix of word vectors ((1) and (3))
and (ii) the focus
is on the convolutional layer of the CNN or the hidden layers
of the RNN ((2) and (4)).
For focus on the input, we apply tanh to the word vectors
(see part (1) of figure) to improve results.

The \emph{source} of attention is the information source
that is used to compute the attention weights, corresponding
to the input of $f$ in \eqref{attweight}.

\eqref{internalatt} formalizes the case in
which focus and source are identical (both are based only on $x$). 
We call this \textbf{internal attention} (see left part of \figref{schemes}).
\emph{An attention layer is called internal if both focus and source are based 
only on information internally available to the network (through input or 
hidden layers).}\footnote{Gates, e.g., the weighting of $h^{t-1}$ in 
\eqref{reset},
can also be viewed as
  internal attention 
mechanisms.}
  
If we conceptualize attention in terms of source
and focus, then a question that arises is
whether we can make it more powerful by \emph{increasing
the scope of the source beyond the input}. 

In this paper, we propose a way of expanding the
source of attention by making an \emph{external resource} $C$
available to the scoring function $f$:
\begin{equation}
 f(x_i) = f'(x_i,C)
 \label{externalattgen}
\end{equation}
We call this \textbf{external attention} (see right part of \figref{schemes}).
\emph{An attention layer is called external if its source
  includes an external resource.}

The specific external-attention scoring function we use for
uncertainty detection is parametrized by $U_1$, $U_2$ and $V$ and defined as follows:
\begin{equation}
 f(x_i) = \sum_j V^T \cdot \tanh(U_1 \cdot x_i + U_2 \cdot
 c_j) \label{externalatt}
\end{equation}
where $c_j$ is a vector representing a cue phrase $j$ of
the training set. We compute $c_j$ as the average of the
embeddings of the constituent words of $j$.

This attention layer scores an input word $x_i$ by
comparing it with each cue vector $c_j$ and summing the
results. The comparison is done using a fully connected
hidden layer. Its weights $U_1$, $U_2$ and $V$ are learned during training.
When using this scoring function in
\eqref{attweight}, each
$\alpha_i$ is an assessment of how important $x_i$ is
for uncertainty detection, taking into account our knowledge
about cue phrases.  Since we use
embeddings to represent words and cues, uncertainty-indicating phrases that did
not occur in training, but are similar to training cue phrases
can also be recognized.

We use this novel attention mechanism for uncertainty detection, but it is
also applicable to other tasks and
domains as long as there 
is a set of vectors available that is analogous to our $c_j$ vectors, i.e.,
vectors that model relevance of embeddings to the task at
hand (for an outlook, see Section \ref{outlook}). 

\subsection{Sequence-agnostic vs.\ Sequence-preserving
Selection}
\seclabel{seqpreserving}
So far, we have explained the basic architecture of an
attention layer: computing attention weights and
reweighting the input. We now turn
to the \emph{integration of the attention layer} into the overall
network architecture, i.e., how it is connected to
downstream components.

The most frequently used downstream connection of the
attention layer is to take the \textbf{average}:
\begin{equation}
a = \sum_i a_i 
\eqlabel{tradAttention}
\end{equation}
We call this the average, not the sum, because
the $\alpha_i$ are normalized to sum to 1 and the standard
term for this is ``weighted
average''.

A variant  is
the \textbf{k-max average}:
\begin{equation*}
a = \sum_{R(\alpha_j)\leq k} a_j
\end{equation*}
where $R(\alpha_j)$ is the rank of $\alpha_j$ in the list of 
activation weights $\alpha_i$
in descending order. This type of averaging is more similar
to k-max pooling and may be more robust
because elements with low weights (which may just be noise)
will be ignored.

Averaging  destroys order
information that may  be needed for NLP sequence classification tasks.
Therefore, we also investigate a sequence-preserving method,
\textbf{k-max sequence}:
\begin{equation}
 a = [a_j | R(\alpha_j)\leq k]
\end{equation}
where $[a_j | P(a_j)]$ denotes the subsequence of 
sequence $A=[a_1,
  \ldots ,a_J]$ from which members not satisfying predicate
$P$ have been removed. Note that sequence $a$ is in the
original order of the input, i.e., not sorted by value.

K-max sequence selects a subsequence of input
vectors. Our last integration method is 
\textbf{k-max pooling}. 
It ranks each
dimension of the 
vectors individually, thus the resulting values
can stem from different input positions. 
This is the same as standard k-max pooling in CNNs 
except that
each vector element in $a_j$ has been weighted (by its
attention weight $\alpha_j$), whereas in standard k-max
pooling it is considered as is. Below, we also refer to k-max
sequence as ``per-pos'' and to k-max pooling as ``per-dim'' to 
clearly
distinguish it from k-max pooling done by the
CNN.

\paragraph{Combination with CNN and RNN Output.}
Another question is whether we combine the attention
result with the result of the convolutional or recurrent layer
of the network.
Since k-max pooling (CNN) and recurrent hidden layers
with gates (RNN) have strengths complementary to attention,
we experiment with concatenating the attention information
to the neural sentence representations.
The final hidden layer then has this form:
\[h = \mbox{tanh}(W_1 a + W_2 r + b)\]
with $r$ being either the CNN pooling result 
or the last hidden state of the RNN
(see \figref{wholeNet}).

\begin{figure}
\centering
 \includegraphics[width=.3\textwidth]{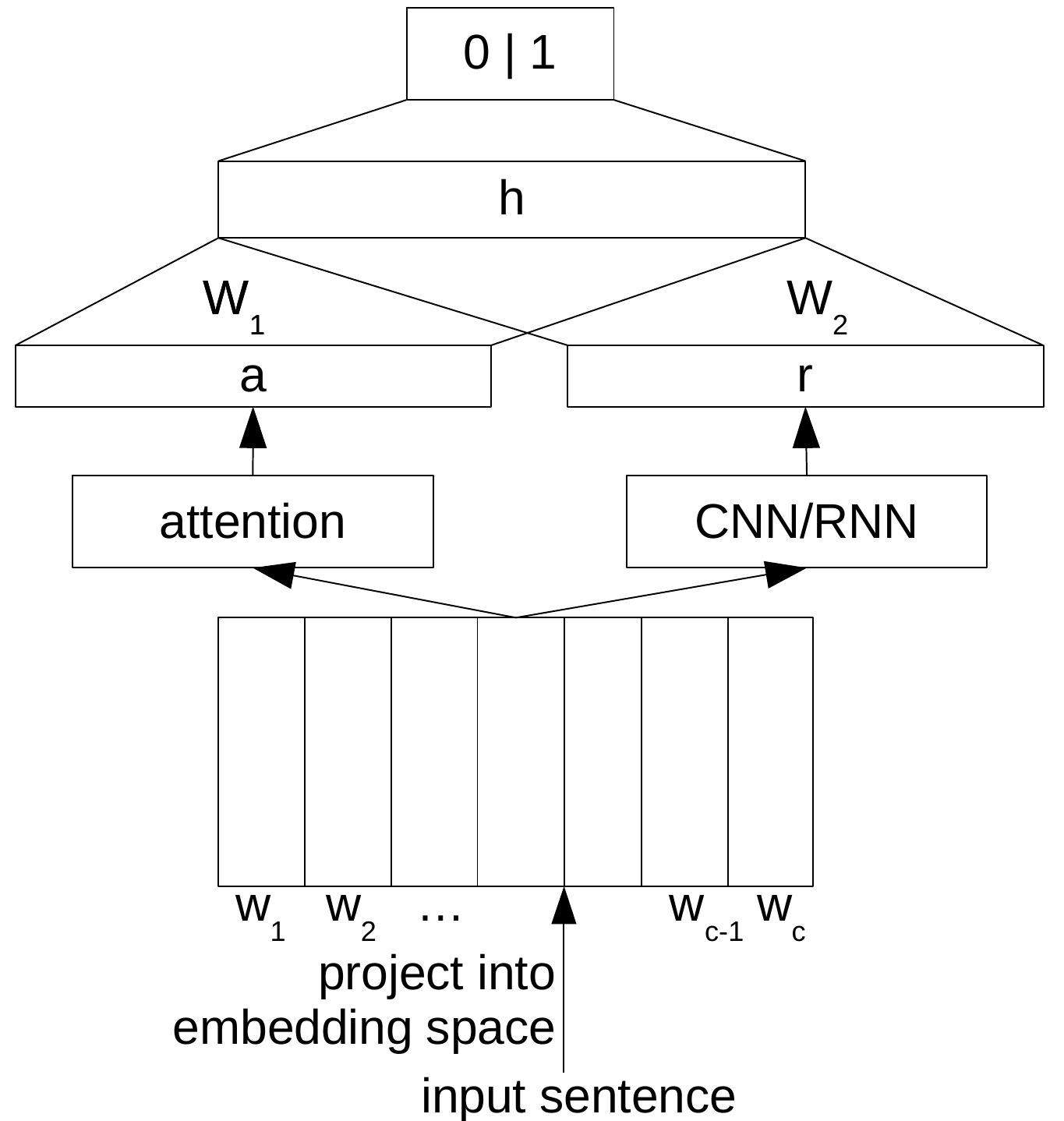}
 \caption{Network overview: combination of attention and CNN/RNN output.
 For details on attention, see \figref{fig-attention}.}
 \figlabel{wholeNet}
\end{figure}

\section{Experimental Setup and Results}
\label{experiments}
\subsection{Task and Setup}
We evaluate on the two corpora of the CoNLL2010 hedge cue detection
task \cite{CoNLLsharedTask}:
Wikipedia  (11,111 sentences in train, 9634 in test) and 
Biomedical (14,541 train, 5003 test).
It is a binary sentence classification task. For each sentence,
the model has to decide whether it contains uncertain
information.

For hyperparameter tuning, we split the training set into 
core-train (80\%) and dev (20\%) sets;
see appendix for
  hyperparameter values.
We
use 400 dimensional word2vec \cite{mikolov}
embeddings, pretrained on Wikipedia, with a special
embedding for unknown words.

For evaluation, we apply the official
shared task measure:  $F_1$ of the uncertain class.

\subsection{Baselines without Attention}
Our baselines are a
support vector machine (SVM) and two standard neural
networks without attention, an RNN and a CNN. 
The SVM is a reimplementation of the top
ranked system on Wikipedia in the CoNLL-2010 shared task
\cite{georgescul}, with parameters set to
\newcite{georgescul}'s values; it uses bag-of-word (BOW)
vectors that only include hedge cues.  Our reimplementation
is slightly better than the published result:
62.01 vs.\ 60.20 on wiki, 78.64 vs.\ 78.50 on bio.

\def\linesvm{1}
\def\linernnzero{2}
\def\linecnnzero{3}
\begin{table}
\centering
\footnotesize
\begin{tabular}{l|l|l|l}
& Model & wiki &  bio\\
\hline
(\linesvm) & Baseline SVM & 62.01$\star$ & 78.64$\star$ \\
(\linernnzero) & Baseline RNN & 59.82$\star$ & \textbf{84.69} \\
(\linecnnzero) & Baseline CNN & \textbf{64.94} & 84.23 \\
\end{tabular}
\caption[$F_1$ results for UD. Baseline models without 
attention.]{$F_1$ results for UD. Baseline models without attention.
$\star$ indicates significantly worse than best model 
(in bold).\protect\footnotemark}
\tablabel{exp1}
\end{table}

\footnotetext{randomization test with $p$$<$.05.}

\begin{table}
\centering
\footnotesize
\begin{tabular}{l|l|l|l}
& Model & wiki &  bio\\
\hline
(\linernnzero) & Baseline RNN & 59.82$\star$ & 84.69 \\
(4) & RNN attention-only & 62.02$\star$ & \textbf{85.32}\\
(5) & RNN combined & 58.96$\star$ & 84.88\\
\hline
(\linecnnzero) & Baseline CNN & 64.94$\star$ & 84.23 \\
(6) & CNN attention-only & 53.44$\star$ & 82.85\\
(7) & CNN combined & \textbf{66.49} & 84.69\\
\end{tabular}
\caption{$F_1$ results for UD. Attention-only vs. combined architectures.
Sequence-agnostic weighted average for attention.
$\star$ indicates significantly worse than best model (bold).}
\tablabel{exp2}
\end{table}

The results of the baselines are given in \tabref{exp1}. 
The CNN (line \linecnnzero) outperforms the SVM (line \linesvm)
on both datasets,
presumably because it considers all words in the sentence --
instead of only predefined hedge cues -- and makes effective
use of this additional information.  The RNN (line \linernnzero)
performs better than the SVM and CNN on biomedical
data, but worse on Wikipedia. 
In Section \ref{CNNvsRNN}, we investigate possible reasons for that.

\subsection{Experiments with Attention Mechanisms}
\label{sec:attentionExperiments}

For the first experiments of this subsection, we
use the sequence-agnostic weighted average for
attention (see \eqref{tradAttention}), the standard in prior work.

\textbf{Attention-only vs.\ Combined Architecture.}
For the case of internal attention, we first remove the 
final pre-output layer of the standard
RNN and the standard CNN to evaluate attention-only architectures.
This architecture works well for RNNs
but not for CNNs.
The CNNs achieve better results when the pooling output (unweighted selection)
is combined with the attention output (weighted selection).
See \tabref{exp2} for $F_1$ scores.

The baseline
RNN has the difficult task of  remembering the entire sentence
over long distances -- the attention mechanism makes this task much
easier. In contrast, the baseline CNN already has an
effective mechanism for focusing on the key parts of the
sentence: k-max pooling. Replacing k-max pooling with
attention decreases the performance in this setup.

Since our main goal is to explore
the benefits of adding attention to existing architectures
(as opposed to developing attention-only architectures), we
keep the standard pre-output layer of RNNs and CNNs in the
remaining experiments and combine it with the attention layer
as in \figref{wholeNet}.

\textbf{Focus and Source of Attention.}
We
distinguish different focuses and sources of
attention.
For focus, we investigate two possibilities: the input to the
network, i.e., word embeddings (F=W); or the 
hidden representations of the RNN or CNN (F=H).
For source,
we compare 
internal (S=I) and external attention (S=E). This gives rise to
four configurations:
(i) internal attention with focus on the first layer of the standard
RNN/CNN (S=I, F=H), see lines (5) and (7) in \tabref{exp2}, (ii) internal attention with
focus on the input (S=I, F=W), (iii) external
attention on the first layer of RNN/CNN (S=E, F=H)
and (iv) external attention on the input (S=E, F=W).
The results are provided in \tabref{exp3}.

For both RNN (8) and CNN (13), the best result is obtained by
focusing attention directly on the word
embeddings.\footnote{The small difference between the RNN
  results on bio on lines (5) and (8) is not significant.}
These results suggest that it is best to optimize
the attention mechanism directly on the input, so that
information can be extracted that is complementary to the
information extracted by a standard RNN/CNN. 

For focus on input (F=W), external
attention (13) is significantly better than internal
attention (11) for CNNs. 
Thus, by designing an architectural element -- external
attention -- that makes it easier to identify hedge cue
properties of words, the learning problem is apparently made
easier. 

For the RNN and F=W, external attention (10) is not better than
internal attention (8): results are roughly tied for bio and
wiki. 
Perhaps the combination of the 
external resource and the more indirect representation
of the entire sentence produced by the RNN is difficult.
In contrast, hedge cue patterns identified by
convolutional filters of the CNN can be evaluated well based on
external attention; e.g., if there is strong
external-attention evidence for uncertainty, then the effect
of a hedge
cue pattern (hypothesized by a convolutional filter) on the 
final decision can be boosted.

In summary, the CNN with external 
attention achieves the best results overall.
It is significantly better than the standard
CNN that uses only pooling, both on Wikipedia and biomedical texts.
This demonstrates that the CNN can make effective use
of external information -- a lexicon of uncertainty
cues in our case.

\begin{table}
\centering
\footnotesize
\begin{tabular}{l|l|l|l|l|l}
& Model & S & F & wiki &  bio\\
\hline
(\linernnzero) & Baseline RNN & - & - & 59.82$\star$ & 84.69 \\
(5) & RNN combined & I & H & 58.96$\star$ & 84.88\\
(8) & RNN combined& I & W & 62.18$\star$ & 84.81\\
(9) & RNN combined& E & H & 61.19$\star$ & 84.62\\
(10) & RNN combined& E & W & 61.87$\star$ & 84.41\\
\hline
(\linecnnzero) & Baseline CNN & - & - & 64.94$\star$ & 84.23$\star$ \\
(7) & CNN combined& I & H & 66.49 & 84.69\\
(11) & CNN combined& I & W & 65.13$\star$ & 84.99\\
(12) & CNN combined& E & H & 64.14$\star$ & 84.73\\
(13) & CNN combined& E & W & \textbf{67.08} & \textbf{85.57}\\
\end{tabular}
\caption{$F_1$ results for UD. Focus (F) and source (S) of attention:
Internal (I) vs external (E) attention; attention on word embeddings (W)
vs. on hidden layers (H). Sequence-agnostic weighted average for attention.
$\star$ indicates significantly worse than best model (bold).}
\tablabel{exp3}
\end{table}

\textbf{Sequence-agnostic vs.\ Sequence-preserving.}
Commonly used attention mechanisms simply
average the vectors in the focus of attention. This means that
sequential information is not preserved. We use
the term sequence-agnostic for this.
In contrast, we propose to investigate sequence-preserving
attention as presented in \secref{seqpreserving}.
We expect this to be important for many NLP tasks.
Sequence-preserving attention is similar to k-max pooling
which also selects an ordered
subset of inputs. While traditional k-max pooling
is unweighted, our sequence-preserving ways of attention
still make use of the attention weights.

\begin{table}
\centering
\small
 \begin{tabular}{l|cc|cc}
&\multicolumn{2}{c|}{average}  & \multicolumn{2}{c}{k-max sequence}  \\
&all & k-max & per-dim & per-pos\\\hline
 Wiki & 67.08 & \textbf{67.52}  & 66.73 & 66.50\\
 Bio & \textbf{85.57} &  84.36 & 84.05 & 84.03
  \end{tabular}
  \caption{$F_1$ results for UD. Model: CNN, S=E, F=W (13). Sequence-agnostic vs. sequence-preserving attention.}
  \tablabel{integration}
\end{table}

\tabref{integration} compares k-max pooling, attention and
two ``hybrid'' designs, as described in \secref{seqpreserving}.
We run these experiments only on
the CNN with external attention focused on word embeddings
(\tabref{exp3}, line 13), the best performing
configuration in the previous experiments.

First, we investigate what happens if we ``discretize''
attention and only consider the values with the top k 
attention weights.
This increases  performance  on wiki
(from 67.08 to 67.52) and decreases it on bio
(from 85.57 to
84.36). We would not expect large differences since
attention values tend to be peaked, so for common values of
k ($k \geq 3$ in most prior work on k-max pooling) we are effectively
comparing two similar weighted averages, one in which most
summands get a weight of 0 (k-max average) and one in which
most summands get weights close to 0 (average over all,
i.e., standard attention).

Next, we compare sequence-agnostic
 with sequence-preserving attention. As described
in \secref{seqpreserving}, two variants are
considered. In k-max pooling, we select the k largest
weighted values per dimension (per-dim in \tabref{integration}). 
In contrast, k-max sequence
(per-pos) selects all values of the k
positions with the highest attention weights. 

In \tabref{integration}, the
sequence-preserving architectures are slightly worse than
standard attention (i.e., sequence-agnostic averaging), 
but not significantly: performance is different by
about half a point.
This shows that k-max sequence and attention
can similarly be used to select a subset of the
information
available, a parallel that has not been highlighted and
investigated in detail before.

Although in this case,
sequence-agnostic attention is better than
sequence-preserving attention, we would not expect this to
be true
for all tasks. 
Our motivation for introducing
sequence-preserving attention was that the semantic meaning
of a sentence can vary depending on where an uncertainty cue
occurs. However, the core of uncertainty detection is
keyword and keyphrase detection; so, the overall sentence
structure might be less important for this task.  For tasks 
with a stronger natural language understanding component, such
as summarization or relation extraction, on the other hand,
we expect sequences of weighted vectors to outperform
averaged vectors. In Section \ref{outlook}, we show that
sequence-preserving attention indeed 
improves results on a sentiment analysis dataset.

\subsection{Comparison to State of the Art}
Table \ref{tab:soa} compares our models with the state of the art
on the uncertainty detection benchmark datasets.
On Wikipedia, our CNN outperforms the state of the art by more than three points.
On bio, the best model
uses a large number of manually designed features
and an exhaustive corpus preprocessing \cite{tang}.
Our models achieve comparable results
without preprocessing or feature engineering.

\begin{table}
\centering
\small
\begin{tabular}{l|c|c}
 Model & wiki & bio\\
 \hline
 SVM \cite{georgescul} & 62.01 & 78.64\\
 HMM \cite{hmm} & 63.97 & 80.15\\
 CRF + ling \cite{tang} & 55.05 & \textbf{86.79}\\
 \hline
 Our CNN with external attention & \textbf{67.52} & 85.57\\
\end{tabular}
\caption{Comparison of our best model with the state of the art}
\label{tab:soa}
\end{table}

\section{Analysis}
\subsection{Analysis of Attention}
In an analysis of examples for which pooling alone (i.e., the
standard CNN) fails, but attention correctly detects
an uncertainty, two patterns emerge. 

In the
first pattern, we find that there are many cues that
have more words than the filter size (which was 3 in our
experiments), e.g., ``it is widely expected'', ``it has also
been suggested''. The convolutional layer of the CNN is
not able to detect phrases longer than the filter size
while for attention there is no such restriction.

The second pattern consists of cues spread over the whole sentence,
e.g., ``Observations of the photosphere of 47 Ursae
Majoris \emph{suggested} that the periodicity \emph{could
  not} be explained by stellar activity, making the planet
interpretation \emph{more likely}'' where we have set
the uncertainty cues that are distributed throughout
the sentence in italics. 
\figref{weightAnalysis2} shows the distribution of
external
attention weights computed by the CNN for this
sentence. The CNN pays the most attention to the three words/phrases
``suggested'', ``not'' and ``more likely'' that correspond
almost perfectly to the true uncertainty
cues. K-max pooling of standard CNNs, on the other hand,
can only select the k maximum values per dimension, i.e.,
it can pick at most k uncertainty cues per dimension.

\begin{figure}[bt]
\centering
\includegraphics[width=.45\textwidth]{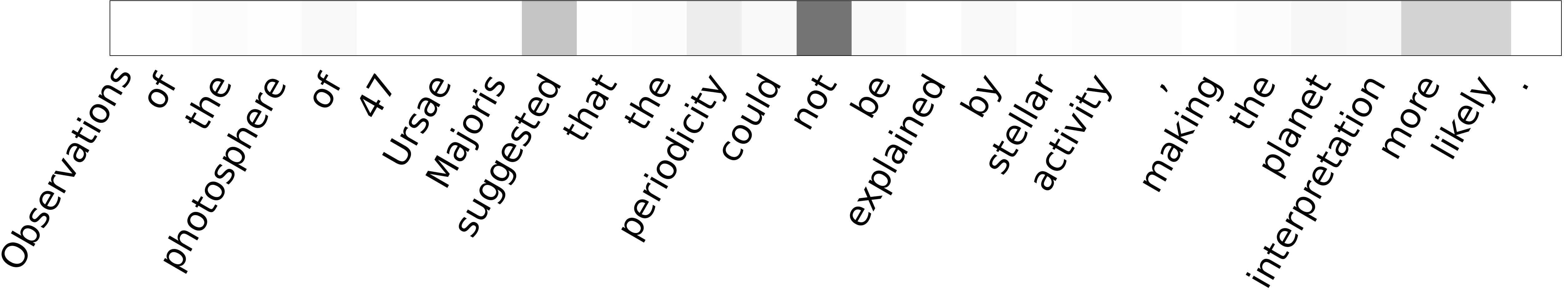}
\caption{Attention weight heat map}
\figlabel{weightAnalysis2}
\end{figure}

\begin{figure}[bt]
\centering
\includegraphics[width=.4\textwidth]{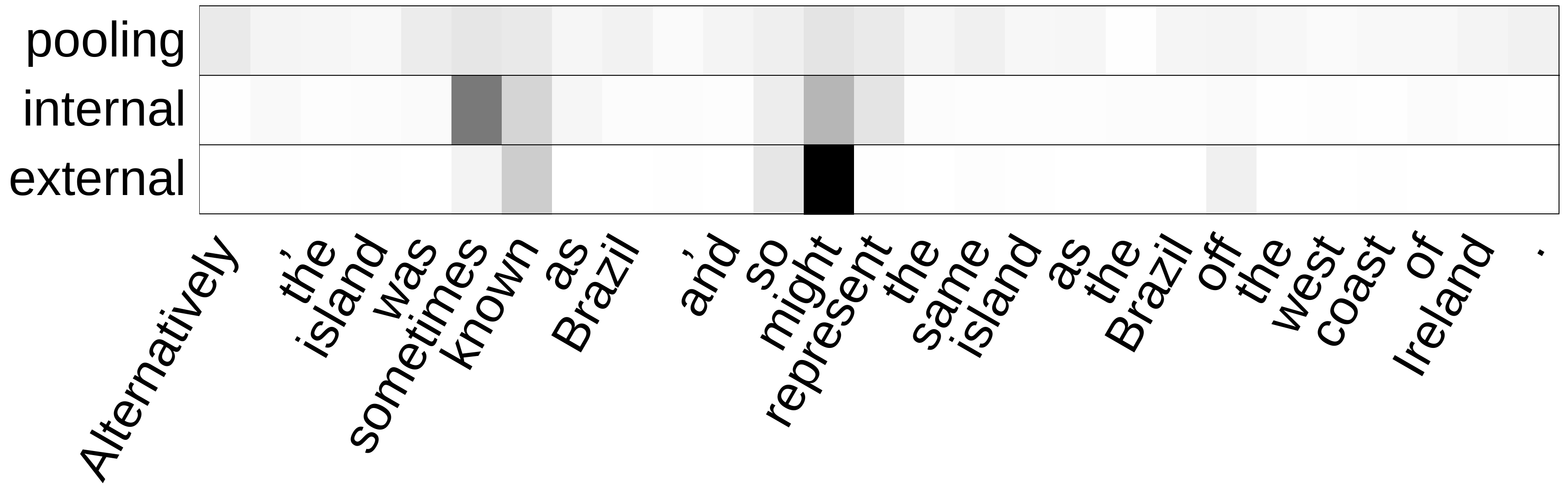}
\caption{Pooling vs. internal vs. ext. attention}
\figlabel{analysis3}
\end{figure}

\paragraph{Pooling vs. Internal  vs. External Attention.}
Finally, we compare the information that  pooling,
internal and  external attention extract.
For pooling, we calculate 
the relative frequency that a value from an n-gram 
centered around a specific word is picked.
For internal and external attention,
we directly plot the attention weights $\alpha_i$. \figref{analysis3}
shows the results of the three mechanisms for an exemplary sentence.
For a sample of randomly selected sentences, we observed similar patterns:
Pooling forwards information from different parts all over the sentence.
It has minor peaks at relevant n-grams (e.g. ``was sometimes known as'' or
``so might represent'') but also at non-relevant parts (e.g. ``Alternatively''
or ``the same island''). There is no clear focus on uncertainty cues.
Internal attention is more focused
on the relevant words. External attention finally has the clearest focus.
(See appendix for more
examples.)

\subsection{Analysis of CNN vs RNN}
\label{CNNvsRNN}
While the results of the CNN and the RNN 
are comparable on bio, the
CNN clearly outperforms the RNN on wiki.
The datasets vary in several aspects, such as average sentence 
lengths (wiki: 21, bio: 27)\footnote{number
of tokens per sentence after tokenization with Stanford tokenizer \cite{stanfordTok}.},
size of vocabularies (wiki: 45.1k, bio: 25.3k),
average number of out-of-vocabulary (OOV) words per sentence
w.r.t. our word embeddings (wiki: 4.5, bio: 6.5), etc. 
All of those features can influence model performance,
especially because of the different way of sentence processing: 
While the RNN merges all information
into a single vector, the CNN 
extracts the most important phrases and ignores all the rest.
In the following, we analyze the behavior of the two models
w.r.t. sentence length and number of OOVs.

\figref{len2f1} shows the $F_1$ scores on Wikipedia
of the CNN and the RNN with external attention for different sentence
lengths. The lengths have been accumulated, i.e., index 0 on the x-axis
includes the scores for all sentences of length $l \in [0,10)$.
Most sentences have lengths $l < 50$. In this range, the CNN performs
better than the RNN but the difference is small. For longer
sentences, however, the CNN clearly outperforms the RNN. This could be
one reason for the better overall performance.

\begin{figure}
\centering
 \includegraphics[width=.4\textwidth]{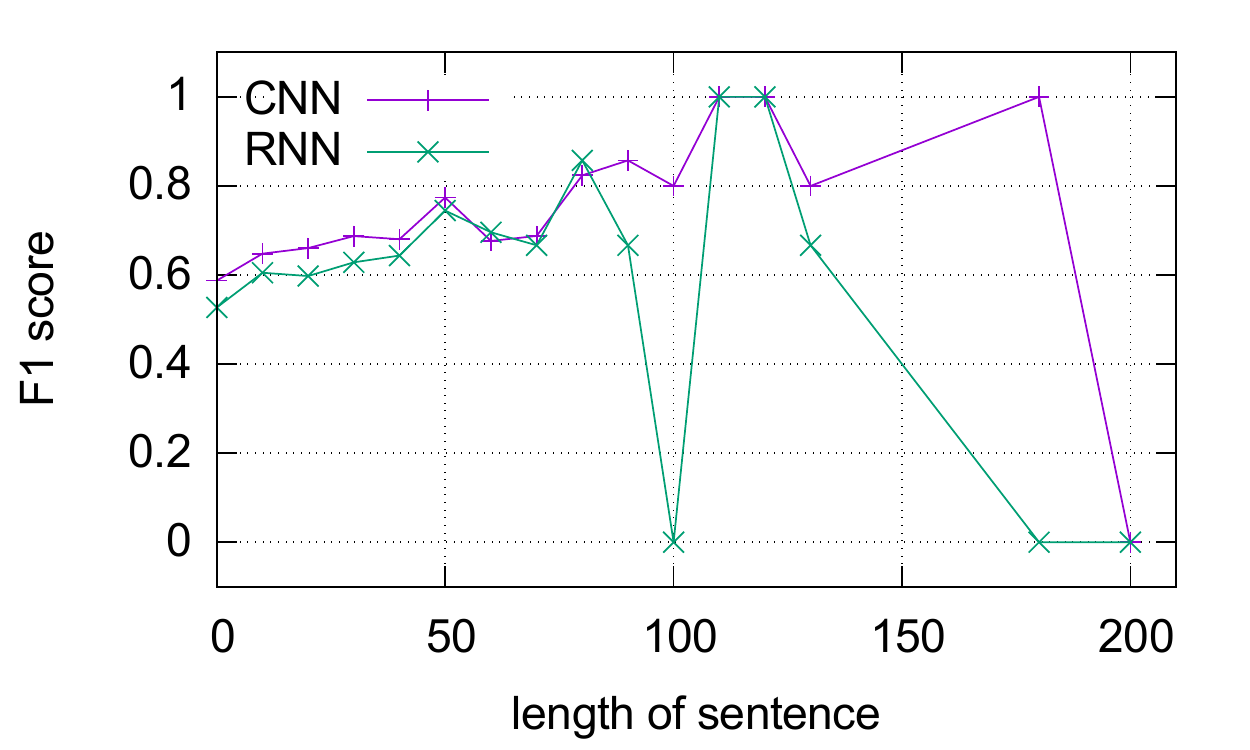}
 \caption{$F_1$ results for different sentence lengths}
 \figlabel{len2f1}
\end{figure}

A similar plot for $F_1$ scores depending on the number
of OOVs per sentence does not give additional insights into the
model behaviors: The CNN performs better than the RNN
independent of the number of OOVs (Figure in appendix).

Another important difference between CNN and RNN is the distribution
of precision and recall. While on bio, precision and recall
are almost equal for both models, the values vary on wiki:
\begin{center}
 \footnotesize
\begin{tabular}{l|l|l}
  & P & R\\
 \hline
 CNN & 52.5 & 85.1\\
 CNN + external attention & 58.6 & 78.3\\
 \hline
 RNN & 75.2 & 49.6\\
 RNN + external attention & 76.3 & 52.0
\end{tabular}
\end{center}
Those values suggest that the RNN predicts uncertainty 
more reluctantly than the CNN.

\section{Outlook: Different Task}
\label{outlook}
To investigate whether our attention methods are also applicable to other tasks,
we evaluate them on the 2-class Stanford Sentiment Treebank (SST-2)
dataset\footnote{\url{http://nlp.stanford.edu/sentiment}} \cite{sst}.
For a baseline model, we train a CNN similar to our uncertainty CNN but with
convolutional filters of different widths, as proposed in \cite{kim2014}, and 
extend it with our attention layer. As cues for external attention,
we use the most frequent positive phrases from the train set. 
Our model is much simpler 
than the state-of-the-art
models for SST-2 but still achieves reasonable 
results.\footnote{The state-of-the-art
accuracy is about 89.5 \cite{sst-soa,mvcnn}.}

\begin{table}
\footnotesize
\centering
 \begin{tabular}{l|l|l|l}
  Model & S & F & test set \\ 
  \hline
  Baseline CNN & - & - & 84.84\\
  CNN attention-only & I & H & 83.56\\
  CNN combined & I & H & 85.22\\
  CNN combined & I & W & 86.11\\
  CNN combined & E & H & 86.06\\
  CNN combined & E & W & \textbf{86.89}
 \end{tabular}
 \caption{Accuracy on SST-2, 
 different focus and source of attention.}
 \label{tab:sentiment1}
\end{table}

\begin{table}
\footnotesize
\centering
 \begin{tabular}{l|l|l|l}
 \multicolumn{2}{c|}{average} & \multicolumn{2}{c}{k-max sequence}\\
  all & k-max & per-dim & per-pos \\ 
  \hline
  86.89 & 86.39 & 87.00 & \textbf{87.22}
 \end{tabular}
 \caption{Accuracy on SST-2, 
 sequence-agnostic vs. sequence-preserving attention.}
 \label{tab:sentiment2}
\end{table}

The results in Table \ref{tab:sentiment1} show the same
trends as the CNN results in \tabref{exp3}, suggesting
that our methods are applicable to other tasks as well.
Table \ref{tab:sentiment2} shows that the
benefit of sequence-preserving attention is indeed
task dependent. For sentiment analysis on SST-2, sequence-preserving
methods outperform the sequence-agnostic ones.

\section{Related Work}
\label{relwork}
\textbf{Uncertainty Detection.}
Uncertainty has been extensively studied in linguistics
and NLP \cite{kiparsky1968,karttunen1973,karttunen},
including modality
\cite{sauri2012,stanfordFactbank,szarvas2012}
and negation \cite{velldal2012,baker2012}. 
\newcite{szarvas2012}, \newcite{vincze2014} and \newcite{crossdomain}
conducted 
cross domain experiments. 
Domains studied include news \cite{factbank},
biomedicine \cite{bioscope}, Wikipedia \cite{weasel} and 
social media \cite{twitter}.
Corpora such as FactBank \cite{factbank} are annotated in
detail with respect to
perspective,  level of factuality and polarity.
\newcite{stanfordFactbank} conducted uncertainty detection
experiments on a version of FactBank extended
by crowd sourcing.
In this work, we use 
CoNLL 2010 shared task data \cite{CoNLLsharedTask} since
CoNLL provides larger train/test sets 
and the CoNLL annotation consists of only two labels (certain/uncertain)
instead of various perspectives and degrees of uncertainty.
When using uncertainty detection for information extraction
tasks like KB population (\secref{intro}),
it is a reasonable first step to consider only two labels.

\textbf{CNNs.}
Several studies showed that CNNs can
handle diverse sentence classification tasks, including
sentiment analysis \cite{kalchbr,kim2014}, relation classification \cite{zeng2014,dosSantos2015}
and paraphrase detection \cite{abcnn}. To our knowledge, we are the first to
apply them to uncertainty detection.

\textbf{RNNs.}
RNNs have mainly been used for sequence labeling or language modeling tasks
with one output after each input token \cite{rnnlm01,rnnlm02}.
Recently, it has been shown that they are also capable of encoding
and restoring relevant information from a whole input sequence.
This makes them applicable to machine translation \cite{gatedRNN,attention0}
and sentence classification tasks \cite{rnnRelClass,attention2,attention3}.
In this study, we apply them to UD for the first
time and compare their results with CNNs.

\textbf{Attention}
 has been mainly used for
recurrent neural networks
 \cite{attention0,attention1,attention2,attention3,attention4,attention5}.
We integrate attention into CNNs
and show that this is beneficial for uncertainty detection.
Few studies in vision integrated attention into CNNs 
\cite{vision01,vision02,vision03}
but this has not been used often in NLP so far.
Exceptions are \newcite{genCNN}, \newcite{attention7} and \newcite{abcnn}.
\newcite{genCNN} used several layers of local and global attention in a
complex machine translation model with a large
number of parameters. Our reimplementation of their network
performed poorly for uncertainty detection
(51.51/66.57 on wiki/bio); we suspect that the
reason is that 
\newcite{genCNN}'s training set was an order of magnitude
larger than ours.
Our approach makes effective use of a much
smaller training set.
\newcite{abcnn} compared attention based input
representations and attention based pooling.
Instead, our goal is to keep the convolutional
and pooling layers unchanged and combine their strengths
with attention.
\newcite{sourceCode} applied a convolutional layer
to compute attention weights. In this work, we
concentrate on the commonly used feed forward layers
for that. Comparing them to other options, such as
convolution, is an interesting direction for future work.

Attention in the literature computes a weighted average
with internal attention weights. In contrast, we
investigate different strategies to incorporate attention
information into a neural network. Also, we propose external
attention.
The underlying  intuition is similar to attention for machine
translation, which learns alignments
between source and target sentences, or attention in question answering,
which computes attention weights based on a question and a fact.
However, these sources for attention are still internal information
of the network (the input or previous output predictions).
Instead, we learn
weights based on
an external source -- a lexicon of cue phrases.

\section{Conclusion}
\label{conclusion}
In this paper, we presented novel attention architectures
for uncertainty detection:
external attention and sequence-preserving attention.
We conducted an extensive set of experiments with
various configurations along different
dimensions of attention, including
different focuses and sources of attention
and sequence-agnostic vs.
sequence-preserving attention.
For our experiments, we used two benchmark
datasets for uncertainty detection and applied
recurrent and convolutional neural networks 
to this task for the first time.
Our CNNs with external attention improved
state of the art by more than 3.5 $F_1$ points on
a Wikipedia benchmark.
Finally,
we showed in an outlook that 
our architectures are applicable to sentiment classification
as well. Investigations of 
other sequence classification tasks are future work.
We made our code publicly available for future research (\url{http://cistern.cis.lmu.de}).

\section*{Acknowledgments}
Heike Adel is a recipient of the Google European Doctoral
Fellowship in Natural Language Processing and this
research is supported by this fellowship.

This work was also supported by DFG (SCHU2246/8-2).

\bibliography{eacl2017}
\bibliographystyle{eacl2017}

\newpage

\appendix

\section{Supplementary Material}

 \subsection{Parameter Tuning}
All parameters and learning rate schedule decisions are based on
results on the development set (20\% of the official training set).
After tuning the hyperparameters (see Tables \ref{tab:params1} and \ref{tab:params2}), 
the networks are
re-trained on the whole training set.

We trained the CNNs with stochastic gradient descent and a fixed learning rate of 0.03.
For the RNNs, we used Adagrad \cite{adagrad} with an initial learning rate of 0.1.
For all models, we used mini-batches of size 10 and applied
L2 regularization with a weight of 1e-5.
To determine the number of training epochs, 
we looked for epochs with peak performances on
the development set.

\begin{table}[h]
\small
\centering
\begin{tabular}{ll||c|c|c|c}
&Model & \# conv & filter & \# hidden & \# att\\
& & filters & width & units & hidden \\
& & & & & units\\
\hline
\hline
     \multirow{5}{*}{\footnotesize\rotatebox{90}{CNN wiki}}& (3) & 200 & 3 & 200 & -\\
 & (6) & 100 & 3 & 500 & -\\
 & (7) & 200 & 3 & 200 & -\\
& (11) & 200 & 3 & 200 & -\\
& (12) & 200 & 3 & 200 & 200\\
 &  (13) & 100 & 3 & 200 & 200 \\
 \hline
     \multirow{5}{*}{\footnotesize\rotatebox{90}{CNN bio}}& (3) & 200 & 3 & 500 & -\\
 & (6) & 100 & 3 & 200 & -\\
 & (7) & 100 & 3 & 500 & -\\
& (11) & 200 & 3 & 200 & -\\
& (12) & 200 & 3 & 500 & 100\\
 &  (13) & 200 & 3 & 50 & 100 
\end{tabular}
\caption{Result of parameter tuning for CNN (``att hidden units'' is the number 
of units in the hidden layer of the attention component);
Model numbers refer to numbers in the main paper
}
\label{tab:params1}
\end{table}

\begin{table}[h]
\small
\centering
\begin{tabular}{ll||c|c|c}
&Model& \# rnn & \# hidden & \# att\\
 & & hidden & units & hidden \\
 & & units & & units\\
\hline
\hline
    \multirow{5}{*}{\footnotesize\rotatebox{90}{RNN wiki}}& (2) & 10 & 100 & -\\
 & (4) & 10 & 100 & -\\
 & (5) &  10 & 200 & -\\
 & (8) &  10 & 100 & -\\
 & (9) & 30 & 200 & 200\\
 & (10) &  10 & 200 & 100\\
\hline
    \multirow{5}{*}{\footnotesize\rotatebox{90}{RNN bio}}& (2) & 10 & 500 & -\\
 & (4) & 10 & 500 & -\\
 & (5) & 10 & 50 & -\\
 & (8) & 10 & 50 & - \\
 & (9) & 30 & 100 & 200\\
 & (10) & 10 & 50 & 200 \\
\end{tabular}
\caption{Result of parameter tuning for RNN}
\label{tab:params2}
\end{table}

\subsection{Additional Examples: Attention Weights}

\begin{figure}[ht!]
\centering
\includegraphics[width=.42\textwidth]{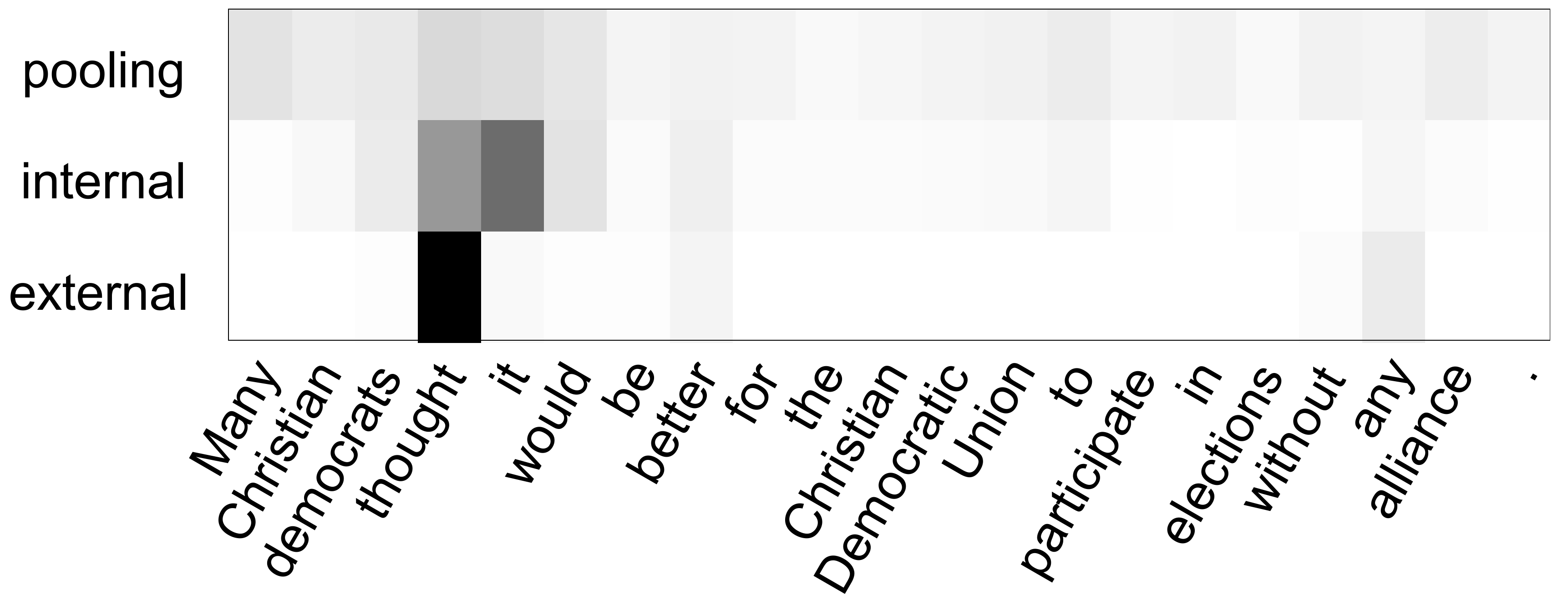}
\caption{Pooling vs.\ internal attention vs.\ external attention}
\figlabel{fig:weightAnalysis3}
\end{figure}

\begin{figure}[ht!]
\centering
\includegraphics[width=.38\textwidth]{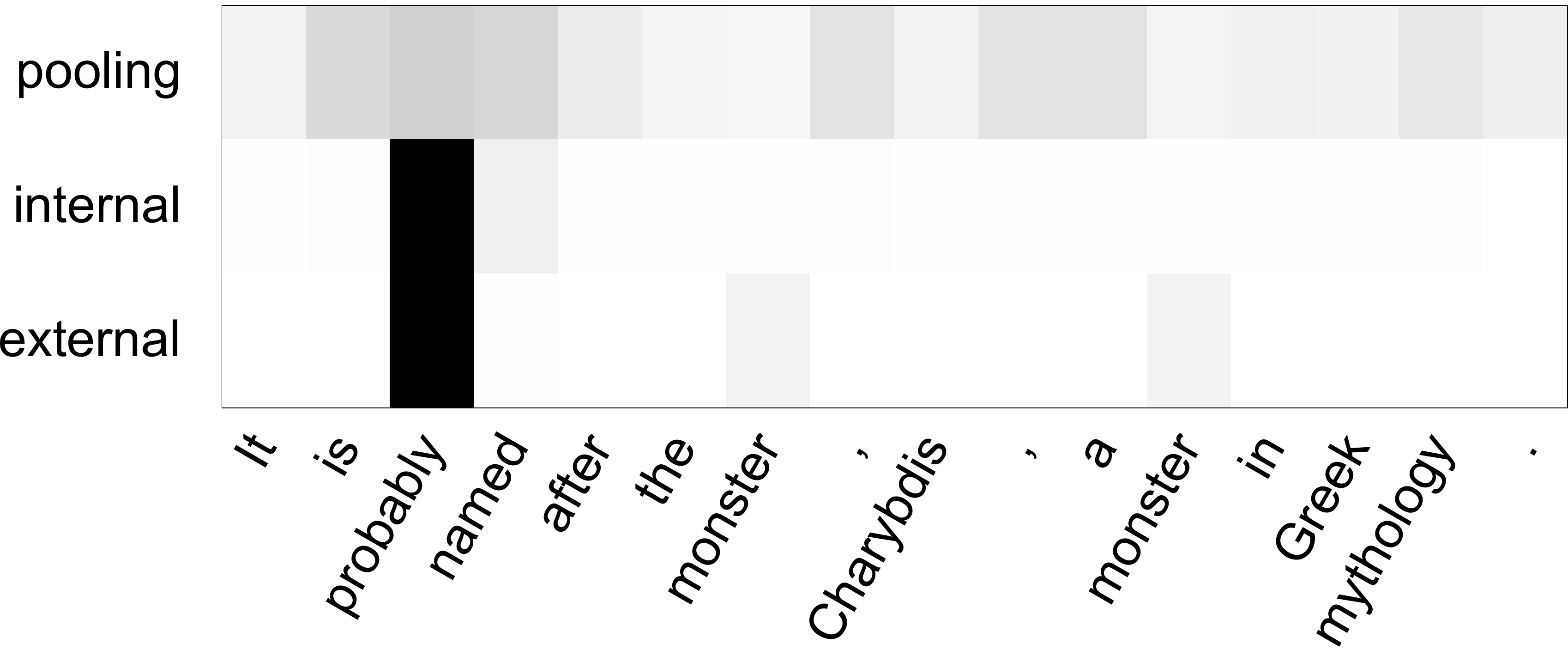}
\caption{Pooling vs.\ internal attention vs.\ external attention}
\figlabel{fig:weightAnalysis4}
\end{figure}

\figref{fig:weightAnalysis3} and \figref{fig:weightAnalysis4}
compare pooling, internal
attention and external attention for randomly picked examples
from the test set.
Again, pooling extracts values from all over the sentence
while internal and external attention learn to focus
on words which can indicate uncertainty (e.g. ``thought'' or
``probably''). 

\subsection{Additional Figure for Analysis: Results Depending on Number of OOVs}
\figref{oov2f1} plots the $F_1$ scores of the CNN and RNN with external
attention w.r.t. the number of out-of-vocabulary (OOV) words 
in the sentences.
The number of OOVs have been accumulated, 
i.e., index 0 on the x-axis includes the score for all sentences with
a number of OOVs in [0,10), etc.
\begin{figure}[h!]
\centering
 \includegraphics[width=.45\textwidth]{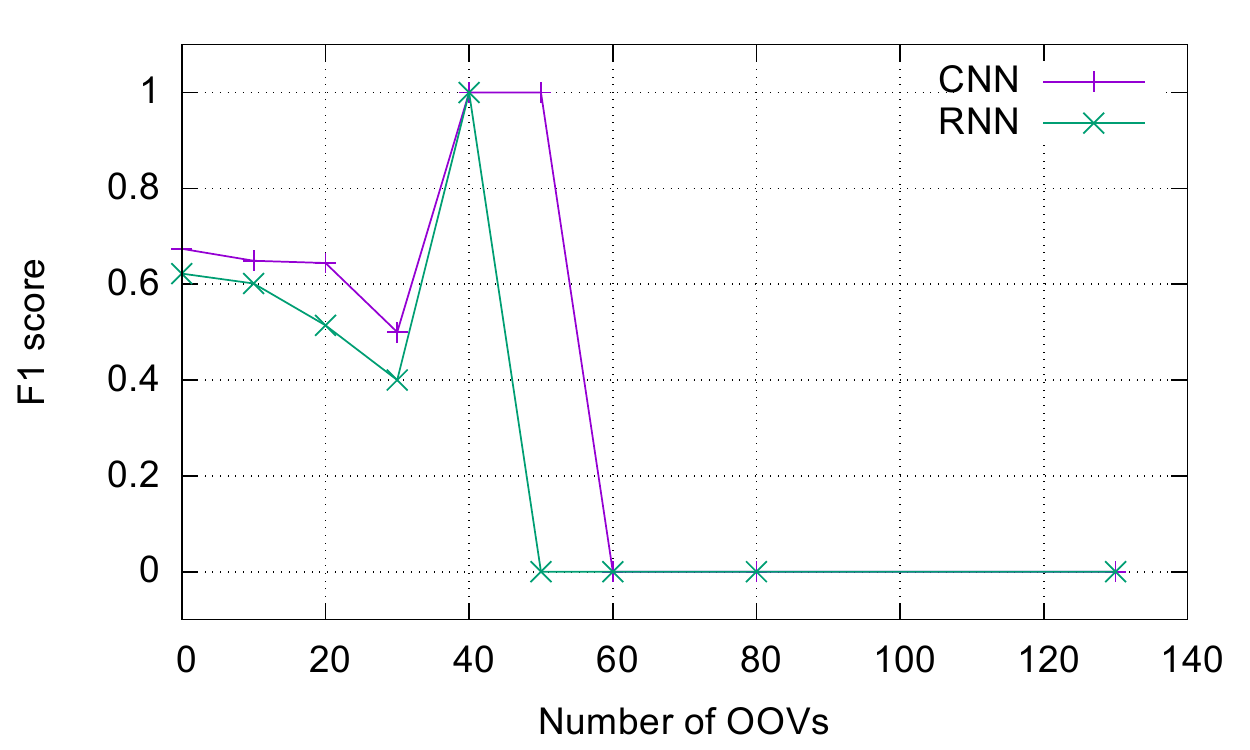}
 \caption{$F_1$ results for different numbers of OOVs in sentence}
 \figlabel{oov2f1}
\end{figure}

\end{document}